\documentclass{article}
\usepackage[utf8]{inputenc}
\usepackage{tikz,amsmath,bm,graphicx,relsize,tikz,subfigure}
\usepackage{amsfonts}
\usepackage[labelfont=bf]{caption}

\usetikzlibrary{bayesnet}
\title{Auxiliary Variables for Multi-Dirichlet Priors}
\author{Christoph Carl Kling\\
GESIS, Cologne\\
datascience@c-kling.de
}
\date{August 2017}

\begin{document}

\maketitle
\begin{abstract}
Bayesian models that mix multiple Dirichlet prior parameters, called \textit{Multi-Dirichlet priors} (MD) in this paper, are gaining popularity~\cite{lin2012coupling,kling2014detecting}.
Inferring mixing weights and parameters of mixed prior distributions seems tricky, as sums over Dirichlet parameters complicate the joint distribution of model parameters.

This paper shows a novel auxiliary variable scheme which helps to simplify the inference for models involving hierarchical MDs and MDPs. Using this scheme, it is easy to derive fully collapsed inference schemes which allow for an efficient inference.
\end{abstract}

\section{Introduction}
In order to develop a collapsed variational inference for hierarchical Dirichlet processes, Teh et al.~\cite{teh08b} introduced an auxiliary variable scheme based on truncated Dirichlet processes.
The same auxiliary variables can be used for efficiently inferring hierarchical models of multinomial distributions with Dirichlet-distributed parameters.

Dirichlet priors can mix multiple parameters to realise predictions of multinomial parameters based on multiple influence factors~\cite{lin2012coupling,kling2014detecting}. In this paper, we call these priors \textit{Multi-Dirichlet priors} (MD). The sampled parameters can serve as parts of Dirichlet priors in a lower level of hierarchical Dirichlet distributions or Dirichlet processes, paving the way for arbitrarily complex hierarchical MD models.

So far, no collapsed variational inference scheme has been proposed for hierarchical MD distributions. In this paper, an auxiliary variable scheme is proposed which allows to directly employ the collapsed variational inference scheme for hierarchical Dirichlet processes given in~\cite{teh08b}.

\section{Multinomial distributions and Dirichlet priors}
\begin{figure}[t!]
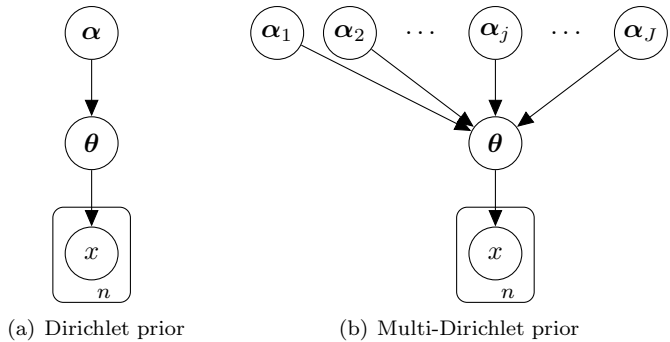

    \centering
        \subfigure[Dirichlet prior]{
        \hspace{1cm}
\tikz{ %
\node[latent] (alpha) {$\boldsymbol{\alpha}$} ; %
\node[latent, below=of alpha,yshift=0.25cm] (p) {$\boldsymbol{\theta}$} ; %
\node[latent, below=of p,yshift=0.25cm] (x) {$x$} ; %

\edge {alpha} {p} ; %
\edge {p} {x} ; %

{\tikzset{plate caption/.append style={below=0pt of #1.south east,yshift=-0.1cm,xshift=-0.2cm}}
 \plate[inner sep=0.15cm,xscale=1,yscale=1,yshift=0.1cm] {plate5} { %
   (x)
} {$n$};}
}
\hspace{1cm}
\label{fig:Dirichlet-prior}
}
\subfigure[Multi-Dirichlet prior]{
\tikz{ %
\node[latent] (alpha) {$\boldsymbol{\alpha}_j$} ; %
\node[left=of alpha, xshift=0.75cm] (dots) {$\dots$} ; %
\node[latent,left=of dots, xshift=0.75cm] (alpha2) {$\boldsymbol{\alpha}_2$} ; %
\node[latent,left=of alpha2, xshift=0.75cm] (alpha1) {$\boldsymbol{\alpha}_1$} ; %
\node[right=of alpha, xshift=-0.75cm] (dots2) {$\dots$} ; %
\node[latent,right=of dots2, xshift=-0.75cm] (alphaJ) {$\boldsymbol{\alpha}_J$} ; %
\node[latent, below=of alpha,yshift=0.25cm] (p) {$\boldsymbol{\theta}$} ; %
\node[latent, below=of p,yshift=0.25cm] (x) {$x$} ; %

\edge {alpha,alpha1,alpha2,alphaJ} {p} ; %
\edge {p} {x} ; %

{\tikzset{plate caption/.append style={below=0pt of #1.south east,yshift=-0.1cm,xshift=-0.2cm}}
 \plate[inner sep=0.15cm,xscale=1,yscale=1,yshift=0.1cm] {plate5} { %
   (x)
} {$n$};}
}
\label{fig:Multi-Dirichlet-prior}
}
    \caption{\textbf{Plate notation for a multinomial distribution with \subref{fig:Dirichlet-prior}~a Dirichlet and \subref{fig:Multi-Dirichlet-prior}~a multi-Dirichlet prior distribution.} The multi-Dirichlet prior is created by summing over parameters $\alpha_{k1}, \dots, \alpha_{jk}$ for each category $k$ of the multinomial.}
    \label{fig:my_label}
\end{figure}

If we sample $n$ single values from a multinomial distribution with parameters $\theta_1, \dots, \theta_K$, the likelihood of observations is is given by
 \begin{align}
     &\operatorname{Mult}(n_1, \dots, n_n \mid \boldsymbol{\theta})
     = \prod_{k=1}^K \theta_k^{n_{k}}
 \end{align}
where $n_k$ are counts which give the frequency of observing the $k$th category in the observations $\boldsymbol{x}$.
The Dirichlet distribution is the conjugate prior distribution of the multinomial:
\begin{align}
    &\operatorname{Dir}\left(\boldsymbol{\theta} \mid \boldsymbol{\alpha}\right) 
    = \frac{\Gamma\left(\sum_{k=1}^K \alpha_{k}\right)}{\prod_{k=1}^K \Gamma\left(\alpha_{k}\right)}
\prod_{k=1}^K
    \theta_k^{ \alpha_{k}-1}
\end{align}
The storyline behind a multinomially distributed variable $x$ with a Dirichlet prior is
 \begin{align}
     \boldsymbol{\theta} \mid \boldsymbol{\alpha} \sim & \operatorname{Dir}\left(\alpha_{1},\dots,\alpha_{K}\right)\nonumber\\
     x_1, \dots, x_n \mid \boldsymbol{\theta} \sim & \operatorname{Mult}\left(\theta_1, \dots, \theta_K\right).
 \end{align}
The graphical model is shown in Fig.~\ref{fig:Dirichlet-prior}.
  
The joint distribution over observations and multinomial parameters of the model is:
 \begin{align}
\operatorname{Mult}(n_{1},\dots,n_{K} \mid \boldsymbol{\theta}) \cdot \operatorname{Dir}(\boldsymbol{\theta} \mid \boldsymbol{\alpha}) = &
\frac{\Gamma(\sum_{k=1}^K \alpha_{k})}{\prod_{k=1}^K \Gamma( \alpha_{k})}
\prod_{k=1}^K
    \theta_k^{ n_{k} + \alpha_{k} -1}.
    \label{eq:dir_mult}
\end{align}
The posterior of a Dirichlet-multinomial distribution is a Dirichlet distribution with parameters $\alpha_1 + n_1, \dots, \alpha_K + n_K$ where the Dirichlet parameters of the prior act as pseudo-counts:
\begin{align}
   \boldsymbol{\theta} \sim \operatorname{Dir}(\alpha_1 + n_1, \dots, \alpha_K + n_K)
\end{align}
Integrating out $\boldsymbol{\theta}$ yields
\begin{align}
&
p(\boldsymbol{n} \mid \boldsymbol{\alpha}) = 
\frac{\Gamma(\sum_{k=1}^K \alpha_{k})}{\Gamma(\sum_{k=1}^K \alpha_{k} + n_{k})}
\prod_{k=1}^K
    \frac{\Gamma(\alpha_{k} + n_{k})}{\Gamma(\alpha_{k})}
    \label{eq:margin_dir_mult}
\end{align}
For efficient inference schemes, auxiliary variables $m_1, \dots, m_K$ can be introduced~\cite{antoniak74,teh08b}:
\begin{align}
& p(\boldsymbol{n},\boldsymbol{m} \mid \boldsymbol{\alpha}) = 
\frac{\Gamma(\sum_{k=1}^K \alpha_{k})}{\Gamma(\sum_{k=1}^K \alpha_{k} + n_{k})}
\prod_{k=1}^K
     \operatorname{s}(m_{k},n_{k}) \cdot \left( \alpha_{k}\right)^{m_k}
     \label{eq:aux-sterling}
\end{align}
where the following equality is used:
\begin{align}
& 
\sum_{m_k = 0}^{n_k} \operatorname{s}(m_{k},n_{k}) \cdot \left( \alpha_{k}\right)^{m_k} 
= \frac{\Gamma(\alpha_{k} + n_{k})}{\Gamma(\alpha_{k})}.
\label{eq:aux_tables}
\end{align}
The auxiliary variables $\boldsymbol{m}$ behave like balls in the Polya urn scheme (or like tables in the Chinese restaurant process for truncated Dirichlet processes)~\cite{teh08b}. The expected values of $\boldsymbol{m}$ are
\begin{align}
    \operatorname{E}\left[ m_k \right] = \alpha_k \cdot \left( \Psi(\alpha_k+n_k) - \Psi(\alpha_k) \right).
\end{align}
If $\alpha_{k}$ contains multinomial parameters, their inference is simple, as $m_1, \dots, m_K$ behave like observed counts of a multinomial. For non-multinomial parameters, further auxiliary variables allow for an elegant inference using gamma distributions~\cite{teh08b}.

\section{The multi-Dirichlet prior}
 Given that we have a multinomial distribution with parameters $\theta_1, \dots, \theta_K$ from which we sample $n$ single values, stored in $x_1, \dots, x_n$. We place a Dirichlet prior over the multinomial parameters $\boldsymbol{\theta}$ which has $K$ parameters. The Dirichlet parameters are calculated by summing over $J$ different $K$-dimensional vectors called \textit{parent prior parameters} $\boldsymbol{\alpha_1}, \dots, \boldsymbol{\alpha_J}$. Then the $k$th parameter of our Dirichlet prior distribution is a sum over parent parameters $\alpha_{1k}, \dots, \alpha_{Jk}$. In the following, this prior distribution will be called a \textit{Multi-Dirichlet} (MD) prior.
 Formally:  
 \begin{align}
     \boldsymbol{\theta} \mid \boldsymbol{\alpha_1}, \dots, \boldsymbol{\alpha_J} \sim & \operatorname{Dir}\left(\sum_{j=1}^J \alpha_{j1},\dots,\sum_{j=1}^J \alpha_{jK}\right)
     ;\quad \alpha_{jk} \in \mathbb{R}_{>0} \, \forall j, k
     \nonumber\\
     x_1, \dots, x_n \mid \boldsymbol{\theta} \sim & \operatorname{Mult}\left(\theta_1, \dots, \theta_K\right)
 \end{align}
 where we show vectors in bold. The graphical model for a multi-Dirichlet distribution is shown in Fig.~\ref{fig:Multi-Dirichlet-prior}.
 
 If we observed counts $n_1, \dots, n_K$ for the $K$ categories from n draws, the joint distribution of observations and parameters becomes:
 \begin{align}
&
\frac{\Gamma(\sum_{k=1}^K \sum_{j=1}^J \alpha_{jk})}{\prod_{k=1}^K \Gamma(\sum_{j=1}^J \alpha_{jk})}
\prod_{k=1}^K
    \theta_k^{ n_{k} + \left(\sum_{j=1}^J \alpha_{jk}\right) -1}
    \label{eq:dir_mult}
\end{align}
The same holds for a Multi-Dirichlet Process (MDP)~\cite{kling2014detecting} with truncation level $K$.

After integrating over the multinomial parameters $\theta$, the joint distribution is:
\begin{align}
p(\boldsymbol{n}\mid \boldsymbol{\alpha_1}, \dots, \boldsymbol{\alpha_J}) = &
\frac{\Gamma(\sum_{k=1}^K \sum_{j=1}^J \alpha_{jk})}{\Gamma(\sum_{k=1}^K \sum_{j=1}^J \alpha_{jk} + n_{k})}
\prod_{k=1}^K
    \frac{\Gamma(\sum_{j=1}^J \alpha_{jk} + n_{k})}{\Gamma(\sum_{j=1}^J \alpha_{jk})}
    \label{eq:margin_dir_mult}
\end{align}

\subsection{Aggregation property}
An important feature of the Dirichlet distribution is the \textit{aggregation property}: Every sum of Dirichlet-distributed parameters follows a Dirichlet distribution where the corresponding parameters were summed up. For our Dirichlet-mul\-ti\-no\-mial distribution we have a Dirichlet-distributed posterior where the parameters are counts plus pseudo counts (the Dirichlet parameters).
\begin{align}
(\theta_{1}, \dots, \theta_{j-1}, \theta_j, \dots, \theta_J) \sim& \operatorname{Dir}(\alpha_1, \dots,  \alpha_{j-1}, \alpha_j,  \dots, \alpha_J)\nonumber\\
\Rightarrow 
(\theta_{1}, \dots, \theta_{j-1} + \theta_j, \dots, \theta_J) \sim& \operatorname{Dir}(\alpha_1, \dots,  \alpha_{j-1} + \alpha_j, \dots, \alpha_J)
\end{align}

Since we have to learn about the contribution of each parent to the Dirichlet prior for learning their parameters, we would like to have individual counts: Instead of storing counts $n_k$ telling us how often we saw category $k$, we would like to store counts $n'_{jk}$ telling how often we saw category $k$ caused by parent $j$. Now if those counts are part of the parameters of a Dirichlet-multinomial posterior, we know that summing over those counts with $\sum_{j=1}^J n'_{jk} = n_k$ yields the original Dirichlet posterior of our model.  
To do so, we introduce probabilities $\theta'_{jk}$ corresponding to the probability of seeing category $k$ explained by the $j$th parent parameter. Then
\begin{align}
   \sum_{j=1}^J n'_{jk} = n_{k}; \qquad  \sum_{j=1}^J \theta'_{jk} = \theta_k
   \label{eq:aux_n_theta}
\end{align}
and the posterior of a multinomial distribution over $\boldsymbol{\theta'}$ with a Dirichlet prior distribution is
\begin{align}
\boldsymbol{\theta'} \sim& \operatorname{Dir}(\alpha_{11} + n'_{11}, \dots,\alpha_{1K} + n'_{1K},\dots, 
\alpha_{J1} + n'_{J1}, \dots, \alpha_{JK} + n'_{JK}
)\nonumber\\
\Rightarrow 
\boldsymbol{\theta} \sim& \operatorname{Dir}(\sum_{j=1}^J \alpha_{j1} + n_{1}, \dots, \sum_{j=1}^J \alpha_{jK} + n_{K})
\end{align}
which directly follows from the aggregation property.

\subsection{Auxiliary variables for category counts}
Based on the aggregation property of the Dirichlet distribution we introduce the auxiliary variables $n'_{jk}$ from Eq.~\ref{eq:aux_n_theta} corresponding to the share of counts assigned to factor $k$ explained by the $j$th parent parameter. We have to account for the possible orderings of parent counts $\boldsymbol{n'}$ and get:
\begin{align}
 &p(\boldsymbol{n},\boldsymbol{n'}\mid \boldsymbol{\alpha_1}, \dots, \boldsymbol{\alpha_J}) =\nonumber
 \\ 
  & \frac{\Gamma(\sum_{k=1}^K \sum_{j=1}^J \alpha_{jk})}{\Gamma(\sum_{k=1}^K \sum_{j=1}^J \alpha_{jk} + n_{k})}
    \prod_{k=1}^K
    {n_{k} \choose n'_{1k}, \dots, n'_{Jk}}
    \prod_{j=1}^J \frac{\Gamma(\alpha_{jk} + n'_{jk})}{\Gamma(\alpha_{jk})}
    \label{eq:disgregation}
\end{align}
which is a product of $K$ Dirichlet-multinomial distributions with prior parameters $\alpha_{1k}, \dots, \alpha_{Jk}$ and therefore
\begin{equation}
    \operatorname{E}[n'_{jk}] = \frac{\alpha_{jk}}{\sum_{j'=1}^J \alpha_{jk'}} \cdot n_{k}.
    \label{eq:exp_aux_counts}
\end{equation}
Summing over all possible values of $n'_{jk}$ yields the original Eq.~\ref{eq:margin_dir_mult}.

In order to get rid of the gamma functions in Eq.~\ref{eq:margin_dir_mult}, Teh et al.~\cite{teh08b} introduced auxiliary variables $m_k$ and $m_{jk}$, corresponding to tables per topic and tables per topic and per parent prior, respectively (see Eq.~\ref{eq:aux_tables}):
\begin{align}
& p(\boldsymbol{n},\boldsymbol{m} \mid \boldsymbol{\alpha_1}, \dots, \boldsymbol{\alpha_J}) =
\frac{\Gamma(\sum_{k=1}^K \sum_{j=1}^J \alpha_{jk})}{\Gamma(\sum_{k=1}^K \sum_{j=1}^J \alpha_{jk} + n_{k})}
\prod_{k=1}^K
     \operatorname{s}(m_{k},n_{k}) \cdot \left(\sum_{j=1}^J \alpha_{jk}\right)^{m_k}
     \label{eq:exp_sum}
\end{align}
where $s(m,n)$ are the unsigned Stirling numbers of the first kind.

\subsection{Auxiliary parent-level counts}
For MD prior distributions, Eq.~\ref{eq:exp_sum} complicates the inference for $\alpha_{jk}$, as the single parts of the parent prior parameters are hidden in exponentiated sums.
However, introducing the parent counts $\boldsymbol{n'}$ from Eq.~\ref{eq:disgregation} and using the auxiliary variables from Eq.~\ref{eq:aux-sterling} we can readily see that
\begin{align}
    &
        p(\boldsymbol{n},\boldsymbol{n'},\boldsymbol{m'}\mid \boldsymbol{\alpha_1}, \dots, \boldsymbol{\alpha_J}) =\nonumber\\
&    \frac{\Gamma(\sum_{k=1}^K \sum_{j=1}^J \alpha_{jk})}{\Gamma(\sum_{k=1}^K \sum_{j=1}^J \alpha_{jk} + n_{k})}
    \prod_{k=1}^K
   {n_k \choose n'_{1k}, \dots, n'_{Jk}}  \prod_{j=1}^J \operatorname{s}(m'_{jk},n'_{jk}) \cdot \alpha_{jk}^{m'_{jk}}
\end{align}
after which the exponentiated sums disappear. The auxiliary variables $\boldsymbol{m'}$ behave like balls in a Polya urn scheme (or like tables in a Chinese restaurant process for truncated Dirichlet processes) where we distinguish not only between $K$ categories, but also distinguish between the $J$ parent prior parameters.

As the choice of parent priors parameters in the Polya urn scheme directly depends on their relative sizes, the expectations for $\boldsymbol{m'}$ are:
\begin{equation}
    \operatorname{E}[m'_{jk}] = \frac{\alpha_{jk}}{\sum_{j'=1}^J \alpha_{j'k}} \cdot m_{k}
    = \alpha_{jk} \cdot \left( \Psi(\alpha_k+n_k) - \Psi(\alpha_k) \right)
    .
    \label{eq:exp_aux_tables}
\end{equation}

If $\alpha_{jk}$ contains multinomial parameters, it now is simple to infer them, as $\boldsymbol{m'}$ behave like observed counts of a multinomial. For non-multinomial parameters, further auxiliary variables allow for an elegant inference using gamma distributions~\cite{teh08b}.

\section{Inference}
For inference on categories, one can directly work with the summed parameters and counts from Eq.~\ref{eq:exp_sum}. These sums can be calculated in advance and can be updated after a fixed number of inference steps.
The resulting inference scheme is identical as for standard Dirichlet-multinomial models~\cite{teh08b}.

For learning about the auxiliary variables for the parent prior parameters, we use the expectations from Eq.~\ref{eq:exp_aux_counts} and Eq.~\ref{eq:exp_aux_tables} which now can be directly calculated given category counts. \textbf{We do not have to explicitly calculate variational distributions over the new auxiliary variables $\boldsymbol{n'}$ and $\boldsymbol{m'}$, which allows for an efficient inference.}
If we separate our parent prior parameters into a mean and a precision part as in~\cite{minka00}, the multinomial mean part of the parent parameters follow a Dirichlet distribution, while the precision part follows a Gamma function, as shown in~\cite{teh08b}. 
\textbf{Using the auxiliary variable scheme, it is also possible to integrate over Dirichlet distributed means of parent priors, which allows for collapsed inference schemes for hierarchical multi-Dirichlet and multi-Dirichlet process models.}

\bibliographystyle{plain}
\bibliography{bibliography}
\end{document}